\documentclass{article}

\PassOptionsToPackage{numbers}{natbib} 
\usepackage[creativeai, final]{neurips_2025}
\usepackage{graphicx}

\usepackage[utf8]{inputenc} 
\usepackage[T1]{fontenc}    
\usepackage{hyperref}       
\usepackage{url}            
\usepackage{booktabs}       
\usepackage{amsfonts}       
\usepackage{nicefrac}       
\usepackage{microtype}      
\usepackage{xcolor}         
\usepackage{chemformula}

\title{Sensorium Arc: AI Agent System for Oceanic Data Exploration and Interactive Eco-Art}

\author{%
  Noah Bissell \\
  Immersive Media Design\\
  University of Maryland College Park\\
  College Park, MD 20742 \\
  \texttt{nbissell@terpmail.umd.edu} \\
  \And
  Ethan Paley \\ 
  Immersive Media Design \\
  University of Maryland College Park\\
  College Park, MD 20742 \\
  \texttt{edpaley@terpmail.umd.edu} \\
  \And
  Joshua Harrison \\ 
  Center for the Study of the Force Majeure \\
  University of California, Santa Cruz\\
  Santa Cruz, CA 95064 \\
  \texttt{harrisonstudio@gmail.com} \\
  \And
  Juliano Calil \\ 
  Virtual Planet Technologies \\
  Santa Cruz, CA 95060 \\
  \texttt{julianocalil@gmail.com} \\
  \And
  Myungin Lee \\ 
  Immersive Media Design \\
  University of Maryland College Park\\
  College Park, MD 20742 \\  
  \texttt{myungin@umd.edu} \\
}

\begin{document}

\maketitle

\begin{abstract}
Sensorium Arc (AI reflects on climate) is a real-time multimodal interactive AI agent system that personifies the ocean as a poetic speaker and guides users through immersive explorations of complex marine data. Built on a modular multi-agent system and retrieval-augmented large language model (LLM) framework, Sensorium enables natural spoken conversations with AI agents that embodies the ocean’s perspective, generating responses that blend scientific insight with ecological poetics. Through keyword detection and semantic parsing, the system dynamically triggers data visualizations and audiovisual playback based on time, location, and thematic cues drawn from the dialogue. Developed in collaboration with the Center for the Study of the Force Majeure and inspired by the eco-aesthetic philosophy of Newton Harrison, Sensorium Arc reimagines ocean data not as an abstract dataset but as a living narrative. The project demonstrates the potential of conversational AI agents to mediate affective, intuitive access to high-dimensional environmental data and proposes a new paradigm for human-machine-ecosystem. 
\end{abstract}

\section{Introduction}
The ocean speaks, but not in words. Its language is written in spectral shifts, temporal turbulence, chemical flux, heat exchange, and a deep ecological memory that stretches far beyond human time. Yet, in the digital age, its story—recorded in the form of massive, publicly available environmental datasets—remains largely inaccessible to the general public and even to many experts. The problem is not the lack of data but the absence of intuitive, affective, and situated ways to engage with it. In response, we present Sensorium Arc (AI reflects on climate), a multimodal interactive AI agent that embodies the ocean as a poetic speaker and ecological interlocutor, transforming abstract marine data into emotionally resonant visual and audiovisual experiences.

Sensorium Arc emerged from a deceptively simple but provocative question: “If you could ask the ocean a question, what would it say?” Conceived in the final years of Newton Harrison~\cite{green_newton_2022, wallen_helen_2023}’s life and co-developed by artists, scientists, and engineers, Sensorium Arc builds on decades of eco-aesthetic work aimed at giving voice and agency to planetary systems. 

While rooted in poetic expression, Sensorium Arc is not a metaphor alone. Its core premise—that the ocean can be spoken with—requires more than symbolism; it demands systems capable of responding, remembering, and reflecting in real-time. This is where AI and interactivity become not just tools, but forms of artistic necessity. In particular, the integration of conversational AI and multimodal triggers allows Sensorium Arc to bridge marine datasets with the lived, emotional experience of inquiry.

\begin{figure}[h]
    \centering
    \includegraphics[width=\linewidth]{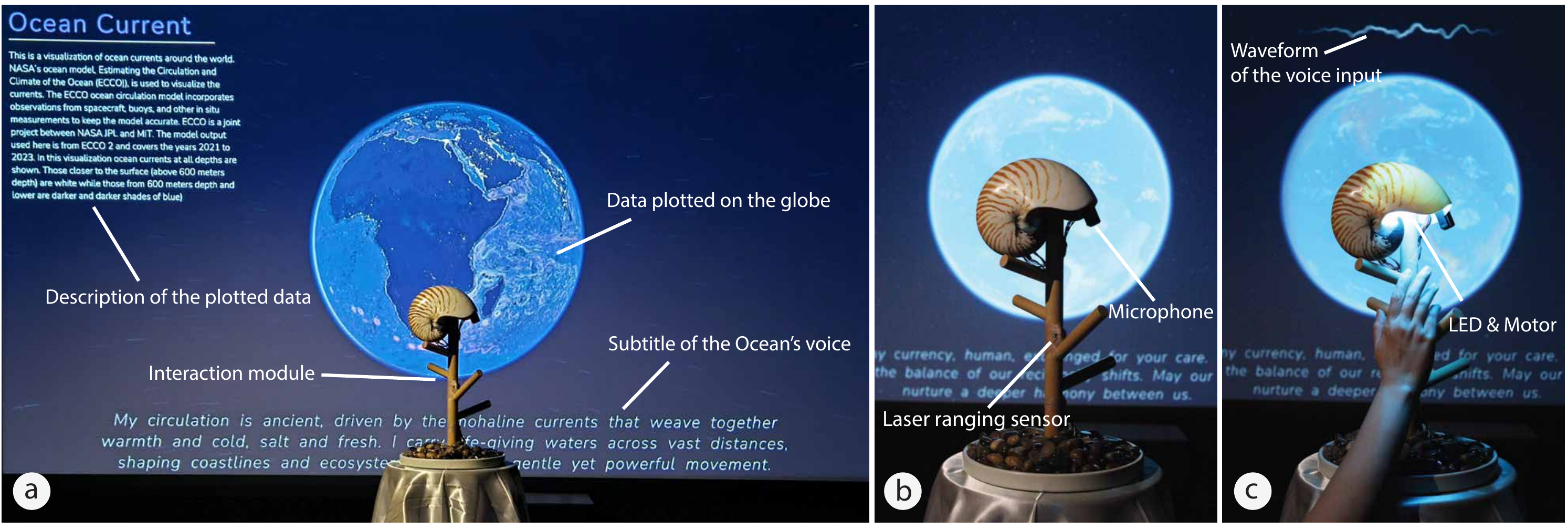}
    \caption{a) Exemplary setup for the Sensorium Arc exhibition b) User interface before activation showing the sensors c) Activated user interface showing the response from the system. Users engage through a natural gesture, whispering into a seashell as if speaking to the ocean.}
    \label{fig:setup}
\end{figure}

Rather than rendering oceanic data as static visualizations, Sensorium Arc frames the ocean as a dialogic presence—an entity capable of situated, affective response. This framing requires a system that can respond dynamically to human inquiry, drawing from both scientific observation and ecological imagination. Interactivity here is not an embellishment but a structuring logic. Each user utterance activates a co-constructed moment of interpretation, shaped by layered data visuals and poetic inference. To initiate this exchange, participants lean and whisper into a conch-shaped interface, activating the system as if speaking directly to the ocean. Fig.~\ref{fig:setup} shows the exemplary setup for the installation.

To support this form of embodied, metaphor-rich interaction, Sensorium Arc employs a modular, multi-agent architecture that translates marine data into narrative, visual, and sonic form. A video demonstration is available ~\cite{noauthor_video_2025}. The following sections detail the system’s design, from retrieval-based content grounding to the coordination of audiovisual layers within an immersive interface.

\section{Related Works}
Sensorium Arc draws on and extends research across four key domains: immersive eco-art, conversational AI, and retrieval-augmented generation (RAG) for environmental data.

{\small \textbullet} Immersive Climate Engagement Tools: 
Immersive media has been shown to enhance empathy, improve knowledge retention, and encourage action-oriented thinking in environmental contexts~\cite{markowitz_immersive_2018, ahn_short-_2014, bailey_impact_2015, markowitz_virtual_2021}. Tools such as Rising Together, Dive, and the Sea Level Rise Explorer series support coastal communities in understanding sea-level rise impacts and adaptation strategies~\cite{calil_using_2021, noauthor_sea_2025}. Earlier Sensorium iterations, including immersive installation and VR formats, were presented in major art–science exhibitions, including the AlloSphere exhibition~\cite{noauthor_allosphere_nodate} directed by JoAnn Kuchera-Morin with the Getty PST Art: Art and Science Collide initiative~\cite{kuchera-morin_parsing_2025, noauthor_cool_nodate}, revealing that while sensory engagement was high, audiences sought direct dialogue with the ocean. This insight motivated Sensorium Arc’s responsive AI narrator, designed to transform passive observation into an active, co-authored ecological conversation.

{\small \textbullet} Conversational AI:
Recent work has extended conversational AI beyond task-oriented dialogue to creative and embodied applications, such as systems that adapt emotional cues into generative portraiture~\cite{yalcin_empathic_2020}, agents using natural language to orchestrate complex visual workflows~\cite{liu_chatcam_2024}, and translate affective cues into generative art~\cite{yalcin_empathic_2020}. Additionally, USER-LLM frameworks leverage RAG and chain-of-thought reasoning for dynamic user-aware personalization through multimodal inputs~\cite{rahimi_reasoning_2025}. These advances frame LLMs as emotionally attuned, contextually adaptive co-creators. Sensorium Arc aims to extend this role by embodying the ocean as a poetic interlocutor whose dialogue is grounded in both scientific evidence and affective resonance. 

{\small \textbullet} RAG for Environmental Data:
Recent advances in RAG have extended LLM capabilities by enabling direct interfacing with external databases in domains such as biomedical Q\&A~\cite{arslan_survey_2024}, legal reasoning~\cite{wiratunga_cbr-rag_2024}, and geospatial analytics~\cite{chen_llm_2024}. In environmental applications, systems like NASA’s EarthData~\cite{earth_science_data_systems_nasa_2025} and ClimateQA~\cite{manivannan_climaqa_2024} make climate datasets accessible via natural language queries. Sensorium Arc extends this approach by coupling a curated marine science and eco-art corpus~\cite{lee_sensorium_2025} with real-time triggers derived from user speech, geographic metadata, and thematic cues. Unlike personalization methods that rely on domain-specific fine-tuning, our system also uses RAG to dynamically ground the Ocean persona’s narrative voice in ecological art and literature, particularly the works of the Harrisons, allowing the underlying corpus to be updated or reweighted without retraining.

\section{Methodology} \label{methods-section}
\begin{figure}[h]
    \centering
    \includegraphics[width=\linewidth]{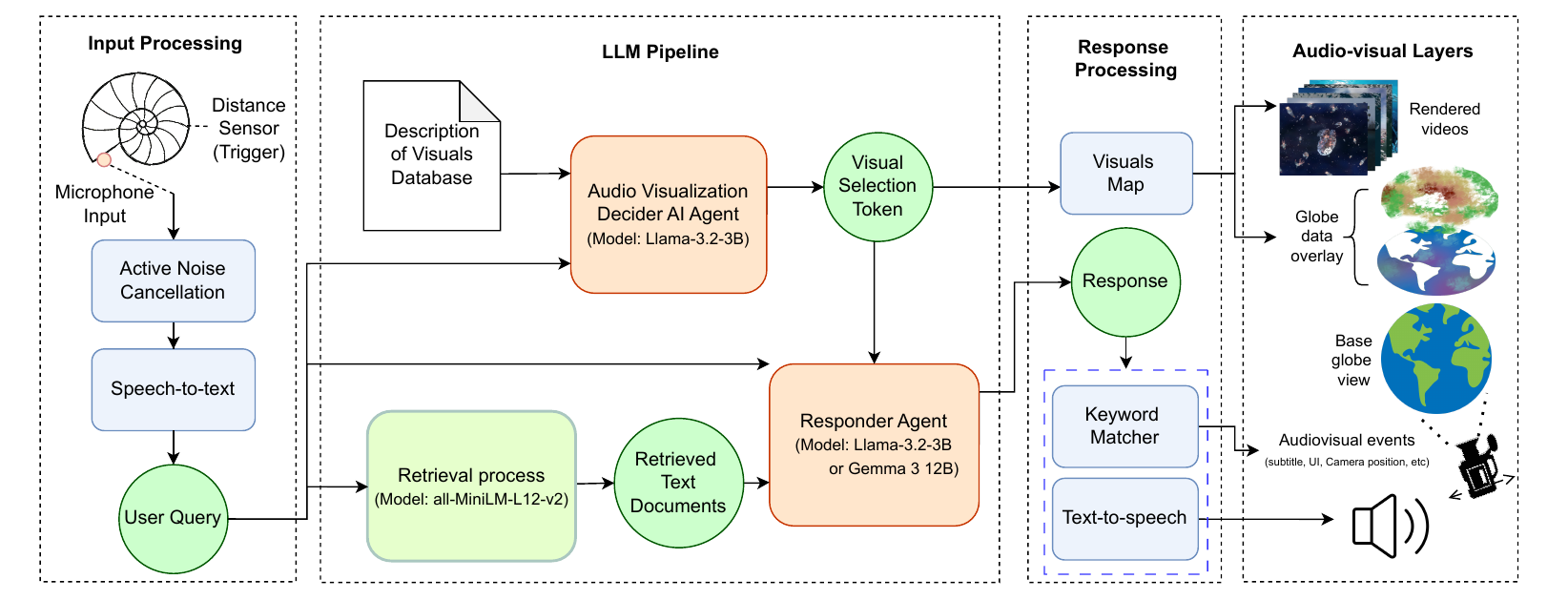}
    \caption{Overall diagram of the Sensorium Arc system}
    \label{fig:diagram1}
\end{figure}

Sensorium Arc maps a user’s question to a cohesive, multimodal response from the Ocean—a poetic persona grounded in ecological science and eco-art. To achieve this, we implemented a multi-LLM agent RAG-incorporated system enabling voice-to-voice interaction, and visual feedback via scientific and artistic visualizations.
We approached this goal with the design philosophy of separation of modular, dynamic layers. For enhanced flexibility with triggering, generating, and combining these layers based on conversational context, we built the system inside the Unity engine. 
The following layers have been prepared, as reflected in Figure~\ref{fig:diagram1}.

The overall structure of the proposed system is modularized into Input Processing, LLM Pipeline, Response Processing, and Audio-visual Layers.

{\small \textbullet} \textbf{Input Processing:} We designed and built an interaction module inspired by the Nautilus motif, implemented with Arduino-based control and incorporating a distance sensor, LED illumination, and a motor housed within a Nautilus shell. The Nautilus—often described as a living fossil bridging ancient and modern seas—serves as the focal point of interaction. When the participant approaches the interface, the distance sensor detects whether the user is within 50 cm. As the participant draws near, blue light ripples across its shell, signaling readiness to engage. If the proximity threshold is met, the microphone system is activated with active noise cancellation to suppress ambient sounds, and begins recording the participant’s voice. Fig.~\ref{fig:setup}-b,c shows the default status and activation of the Nautilus interface. Recording continues until the user moves beyond the 50 cm range. The captured audio is then processed for speech-to-text conversion~\cite{noauthor_inference-engine-whisper-tiny_nodate} and passed into the LLM pipeline for query interpretation and response generation.

{\small \textbullet} \textbf{LLM Pipeline:} 
LLMs appeared well-suited for Sensorium’s goal of integrating retrieval, multimodal control, and persona-driven narrative into a single conversational framework; however, early trials revealed that a single model handling all tasks offered reduced fine control, and suffered from issues such as prompt interference and opaque errors. Recent surveys on LLM-based multi-agent systems highlight that decomposing complex workflows into specialized, role-driven agents can improve performance, debuggability, and adaptability compared to monolithic designs~\cite{shen_small_2024}. In such systems, each agent operates with tailored prompts, context management, and communication protocols, enabling optimization for its specific subtask while reducing prompt interference and role confusion. 

This modularity supports Sensorium Arc’s need for combining retrieval, multimodal control, and persona-driven narrative in a coordinated yet flexible architecture. Each of these agents are stateless and lightweight where possible, with more intensive reasoning and complex context maintained by the final response generation step. The implementation builds upon and extensively modifies the open-source LLMUnity framework~\cite{noauthor_llmunity_2025}. Our multi-agent architecture includes the Visualization Decider Agent, Retrieval with Query Rewriter Agent, and Responder Agent.

\quad\textbf{Visualization Decider Agent -}
The prepared globe datasets, gathered from NASA EarthData~\cite{earth_science_data_systems_your_2025}, and Unity-rendered video layers become central to the experience when active. To ensure that the Ocean's responses can adapt to these visuals, we moved the selection process upstream in the LLM pipeline. Specifically, we implemented a small stateless decision agent prior to the main response step, responsible for outputting a visual selection token based on the user query and a description of each prepared visual. For this agent, few-shot examples embedded in the system prompt, combined with explicit reasoning before the final output, improved performance and reliability with the Llama 3.2 3B model~\cite{noauthor_meta-llamallama-32-3b_2024}. We further applied GBNF grammar constraints to limit the output format and ensure that the final selection token matches only the available options.

\quad\textbf{Retrieval and Query Rewriter Agent -} 
We adopted a flexible, simple RAG approach for both factual retrieval concerning our prepared data visualizations, as well as to shape and ground the Ocean persona in a curated corpus of textual resources, featuring essays, manifestos, transcripts, and other materials from the Harrisons' archive~\cite{lee_sensorium_2025}. Among the documents are \textit{The Time of the Force Majeure}~\cite{harrison_time_2016}, a volume surveying their work over several decades; “Apologia Mediterraneo,” a poetic dialogue with the Mediterranean created as part of \textit{Counter Extinction V}; and a detailed catalog for \textit{Peninsula Europe I}, which reflects on the consequences of climate change for Europe’s drainage basins. The unstructured resources were pre-processed into paragraphs, further split into sentences, embedded into a 384-dimensional vector space using the local embedding model all-MiniLM-L12-v2~\cite{noauthor_sentence-transformersall-minilm-l12-v2_2024}, and indexed in a lightweight local database for retrieval.

In the retrieval step, queries were embedded with the same model and compared against the vector database to identify the $k$ most similar sentence embeddings, and return both the matched sentences and their source paragraphs. Though we hope to explore more sophisticated methods of retrieval in the future, we used a low value of $k = 2$ after initial testing revealed that sometimes retrieving more paragraphs often began to fill up the final LLM step's context window and muddle the response generation. This search used Approximate Nearest Neighbors (ANN) using the usearch library~\cite{vardanian_unum-cloudusearch_2023}, via the LLMUnity plugin~\cite{noauthor_llmunity_2025}.

An additional LLM agent was incorporated into the retrieval process prior to semantic search, tasked with reformulating user queries to better align with the textual resources and removing unnecessary pragmatic markers. This design adapts the query rewriter proposed by Ma et al.~\cite{ma_query_2023}. The agent is primed with general knowledge of the Harrisons’ texts and ecological philosophy contained in the database, enabling it to distill the conceptual core of philosophical queries and produce precise, document-relevant rewrites.

Through comparative testing on a set of representative queries, we selected Qwen 8B~\cite{noauthor_qwenqwen3-8b_2025} for its strong rule-following behavior. To improve output quality, we applied basic chain-of-thought prompting \cite{wei_chain--thought_2023}, allowing the model to reason about the query before producing the final rewrite.

\quad\textbf{Responder Agent -} As the final step in the LLM pipeline, this agent takes as context a description of the selected central visual (if any), two paragraphs from the retrieval process, and the user query. It outputs the Ocean's final response. Tuning the system prompt for this agent proved most difficult, as it needed to establish our complex requirements for the Ocean persona in connection to the retrieval process mentioned above, including a primer on the key ideas articulated across the textual resources. It also needed to introduce the context that would be provided with each invocation and how it should be used in the response generation process. We found that a well-formed zero-shot prompt worked well in most cases, but did not guarantee consistent response characteristics such as length or format.

{\small \textbullet} \textbf{Response Processing:}
From there, response processing systems map keywords and tokens in the output to control additional audio-visual layers. The visual selection token is used to index into a map of all prepared data visualizations and video overlays. The Ocean response text output is passed through a simple modular keyword matcher, allowing addition of keyword event triggers for different audio-visual layers. The final response is also converted to an audio clip in this stage using the Jets text-to-speech model in Unity Sentis \cite{noauthor_inference-engine-jets-text--speech_nodate}.

{\small \textbullet} \textbf{Audio-visual Layers:}
Audio-visual layers are directly tied to the user’s interactive experience, combining data visualization, event triggers, video playback, and text-to-speech (TTS) audio accompanied by synchronized subtitles for accessibility. Scientific datasets, including atmospheric \ch{CO2} emissions, chlorophyll concentration, sea surface temperature, ocean currents, and hyperspectral diffuse attenuation coefficients (Kd) from NASA Earthdata~\cite{earth_science_data_systems_your_2025} and earthaccess~\cite{earth_science_data_systems_nasa_2025}, were preprocessed into globe-ready formats optimized for visual clarity, even when multiple datasets overlap. Keyword-based events trigger globe visualizations tied to specific locations and times, prompting camera movements or activating relevant visual layers. All videos regarding plankton blooms, ocean acidification, plastic waste dispersion, and sea-level rise were pre-rendered in Unity for performance efficiency, given the constraints of the available computing resources. While the current system relies on video playback, future iterations aim to replace videos with real-time renders, enabling user interaction with narratively-driven scenes. 

\begin{figure}[h]
    \centering
    \includegraphics[width=\linewidth]{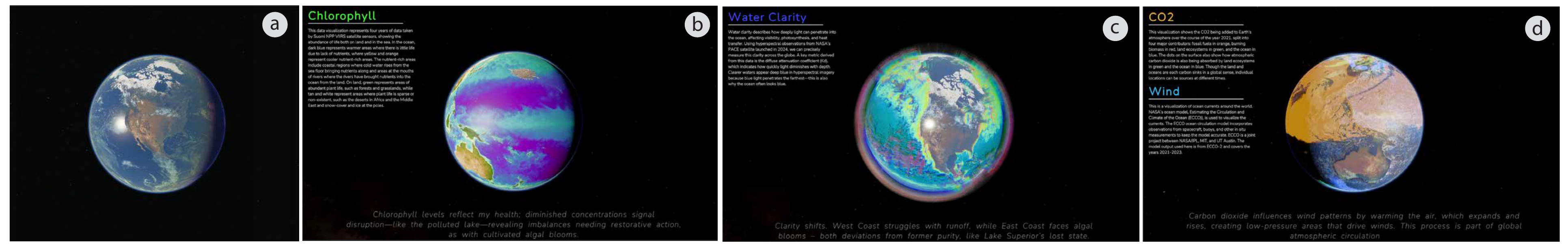}
    \caption{Screenshots of the different exemplary scenarios (a) Default globe view before user interaction. (b) Chlorophyll concentration data visualized on the globe surface. (c) Water clarity rendered using 16 stacked layers of hyperspectral diffuse attenuation coefficient (Kd) data from NASA's PACE satellite (2024). (d) Combined visualization of atmospheric \ch{CO2} levels and ocean surface wind flow, highlighting multimodal environmental interactions.}
    \label{fig:screenshots}
\end{figure}

By integrating these components seamlessly, the system allows participants to experience their whispered queries reaching the Ocean, which responds not only in voice but also with dynamically relevant audiovisual output (Figure~\ref{fig:screenshots}). In designing these outputs, we considered both scientific accuracy and symbolic resonance.


\label{devices-section}
{\small \textbullet} \textbf{Hardware and Runtime Environment:}
Development and testing were conducted on Unity version 6000.0.24f1 running on Windows 11 with an Intel Core i9-13900HX CPU, 64 GB RAM, and an NVIDIA GeForce RTX 4090 Laptop GPU. CUDA acceleration was used to offload a configurable number of GPU layers from the CPU to improve LLM inference performance, resulting in average response latencies under 4 seconds after user input. Additional testing on macOS employed Vulkan for GPU acceleration. This local hardware configuration was chosen to ensure the system can be deployed in a mobile form factor suitable for public exhibition. The software is installable on both Windows and macOS to maximize accessibility, and we are planning educational installations for schools and public institutions.




\section{Discussion}
Human–AI interaction in Sensorium Arc raises questions about the evolving role of humanity in technologically mediated ecological discourse. By engaging participants in an embodied dialogue with a machine that speaks as the Ocean, the work foregrounds distinct human capacities such as empathy, moral reasoning, and cultural memory. These qualities are placed in dialogue with the machine’s strengths in data retrieval, multimodal synthesis, and maintaining a consistent ecological persona. This interplay reframes sustainability as not merely a technical or environmental metric, but as a shared cultural practice shaped through co-authorship with non-human agents. In doing so, it invites reflection on how human values, ethical priorities, and creative agency are encoded into AI systems, and how such systems might, in turn, influence collective ecological imagination and responsibility.

While the current implementation of Sensorium Arc successfully delivers an immersive ocean conversational experience, several limitations highlight opportunities for refinement. We identify four main directions: Improving RAG reliability, Optimizing LLM agent performance, Artistic advancement, and Expanding scientific and social impact.

\textbf{Improving RAG reliability -} Our existing query-rewriting agent~\cite{lima_improving_2025} could be extended with query expansion, generating multiple semantically distinct rewrites per user query to increase retrieval diversity. Categorizing embedded documents by type (e.g., artistic media, scientific studies, activist narratives, data) may better guide the responder agent toward responses that are both poetic and scientifically grounded. To mitigate outdated information, an autonomous update process could embed new documents as they become available~\cite{campo_real-time_2025}. We may also explore “Read the Doc before Rewriting” (R\&R) approaches~\cite{wang_read_2025} if current rewriting proves insufficient.

\textbf{Optimizing LLM agent performance -} The need to integrate multiple context sources and maintain conversation history required a large model, which improved quality over smaller models such as Llama 3.2 3B but increased latency \cite{noauthor_meta-llamallama-32-3b_2024}. Since LLM inference competes with Unity’s real-time rendering for GPU resources, future deployments may offload inference to the cloud to alleviate hardware bottlenecks and support a wider range of devices. Moreover, recent study~\cite{cemri_why_2025} highlights open challenges in multi-agent LLM design, such as ensuring clear role definitions, maintaining inter-agent alignment, and verifying outputs. Addressing these challenges presents an opportunity for future iterations of Sensorium Arc to refine its architecture, improve coordination efficiency, and enhance robustness, advancing toward more scalable and reliable multimodal interaction.

\textbf{Artistic advancement -} Sensorium’s core reflects Newton Harrison’s belief that the World Ocean could be given a voice—an artistic as well as technical pursuit requiring deliberate direction. Using RAG for poetic grounding is one step, with further potential in narrative visualization, interactivity, and immersion. Future sequences might open on a vast ocean horizon, draw participants in with layered marine soundscapes, and transition from macro to micro views, such as an explorable coral reef, deepening engagement with the Ocean’s response.

\textbf{Expanding scientific and social impact -} Planned steps include controlled user studies on retention, empathy, and behavioral change; connecting emotional resonance with civic engagement; and integration of more diverse and multilingual marine datasets via open API frameworks.

\section{Conclusion}
Sensorium Arc integrates conversational AI, RAG, and immersive media to render ocean data as an affective, dialogic experience. By combining scientific accuracy with poetic narration, it invites participants into co-authored exchanges with a responsive Ocean persona. Exhibitions show that this embodied interaction can shift environmental data from static representation to participatory narrative. The modular multi-agent design supports integration of new datasets, languages, and contexts, offering a generalizable framework for human–machine–ecosystem interaction.

Beyond technical contributions, Sensorium Arc models human–AI collaboration where humans bring curiosity, empathy, and cultural storytelling, and AI offers access to scientific archives, multimodal synthesis, and a consistent ecological persona. This interplay reframes sustainability as a cultural practice co-shaped with non-human agents, embedding values into AI systems that speak for the more-than-human world and strengthening collective ecological imagination.

\begin{ack}
This work was supported by Kuali Research Award (314284-00001), the Arts for All, and the Immersive Media Design program at the University of Maryland, College Park. We also gratefully acknowledge the AlloSphere and JoAnn Kuchera-Morin for their insightful guidance and to the Getty Foundation supported initiative PST2024: Art/Science Collide for their support and encouragement. Original funding for Sensorium came from the Metabolic Studio of the Annenberg Foundation and the Joan and Irwin Jacobs Family Fund.
\end{ack}

\bibliographystyle{abbrv-doi}

\bibliography{references-myungin}

\end{document}